\pdfoutput=1
\documentclass[10pt,twocolumn,letterpaper]{article}
\usepackage{cvpr}
\usepackage{times}
\usepackage{epsfig}
\usepackage{graphicx}
\usepackage{amsmath}
\usepackage{amssymb}

\usepackage{algorithm}
\usepackage{algorithmic}
\usepackage{caption}

\usepackage[pagebackref=true,breaklinks=true,letterpaper=true,colorlinks,bookmarks=false]{hyperref}

\cvprfinalcopy 

\def\cvprPaperID{813} 

\begin{document}

\title{Generalized Video Deblurring for Dynamic Scenes}

\author{Tae Hyun Kim and Kyoung Mu Lee\\
	Department of ECE, ASRI, Seoul National University, 151-742, Seoul, Korea\\
	{\tt\small \{lliger9, kyoungmu\}@snu.ac.kr, http://cv.snu.ac.kr}
}

\twocolumn[{%
	\renewcommand\twocolumn[1][]{#1}%
	\maketitle
	\begin{center}
		\centering
		\includegraphics[width=\textwidth]{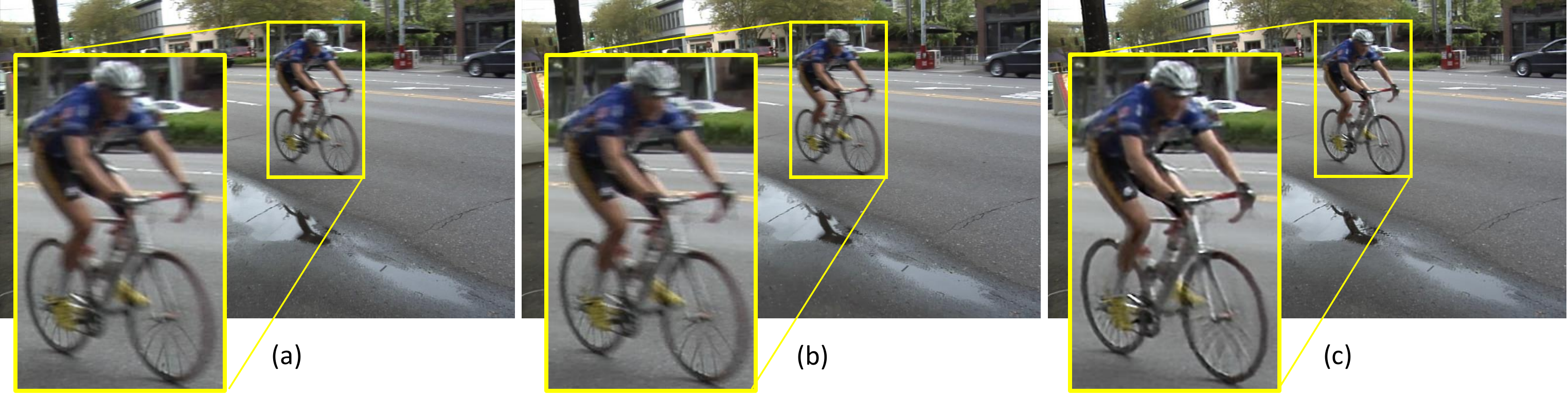}
		\captionof{figure}{(a) Blurry frame of the bicycle sequence. (b) Deblurring result of Cho et al.\cite{cho_siggraph2012}. (c) Our result.}
		\label{fig_teaser}
	\end{center}%
}]

\maketitle
\thispagestyle{empty}

\begin{abstract}
Several state-of-the-art video deblurring methods are based on a strong assumption that the captured scenes are static. 
These methods fail to deblur blurry videos in dynamic scenes. 
We propose a video deblurring method to deal with general blurs inherent in dynamic scenes, contrary to other methods. 
To handle locally varying and general blurs caused by various sources, such as camera shake, moving objects, 
and depth variation in a scene, 
we approximate pixel-wise kernel with bidirectional optical flows. 
Therefore, we propose a single energy model that simultaneously estimates optical flows 
and latent frames to solve our deblurring problem. 
We also provide a framework and efficient solvers to optimize the energy model. 
By minimizing the proposed energy function, we achieve significant improvements in removing blurs 
and estimating accurate optical flows in blurry frames. 
Extensive experimental results demonstrate the superiority of the proposed method in real 
and challenging videos that state-of-the-art methods fail in either deblurring or optical flow estimation.

\end{abstract}

\section{Introduction}
Motion blurs are the most common artifacts in videos recorded using hand-held cameras.
For decades, several researchers have studied deblurring algorithms to remove motion blurs.
Their methodologies depend on whether the captured scenes are static or non-static.
Early works on single image deblurring usually assumed that the scene is static with constant depth~\cite{Cho:2009,Fergus:2006,Gupta:2010,Hirsch:2011,Shan:2008,Whyte:2012}.
The successful approaches were naturally extended to video deblurring. 
In the work of Cai et al.~\cite{cai2009blind}, a multi-image deconvolution method was proposed
using sparsity of blur kernels and clear image to handle registration errors.
However, this method only enables two-dimensional translational camera motion, which generates uniform blur.
Therefore, the proposed approach cannot handle rotational camera shake,
which is the main cause of large motion blurs~\cite{Whyte:2012}.
To overcome this limitation, Li et al.~\cite{li2010generating} used a method 
parameterizing spatially varying motions with 3x3 homographies,
and could handle spatially varying blurs by camera rotation.
In the work of Cho et al.~\cite{cho2012registration}, camera motion in three-dimensional space was estimated
without the assistance of specialized hardware.
In addition, non-uniform blurs by projective camera motion could be removed.
Spatially varying blurs by depth variation in a static scene was handled recently
in the works of Lee and Lee~\cite{lee2013dense} and Paramanand et al.~\cite{paramanand2013non}.

\begin{figure*}[t]
	\begin{center}
		\includegraphics[width=\linewidth]{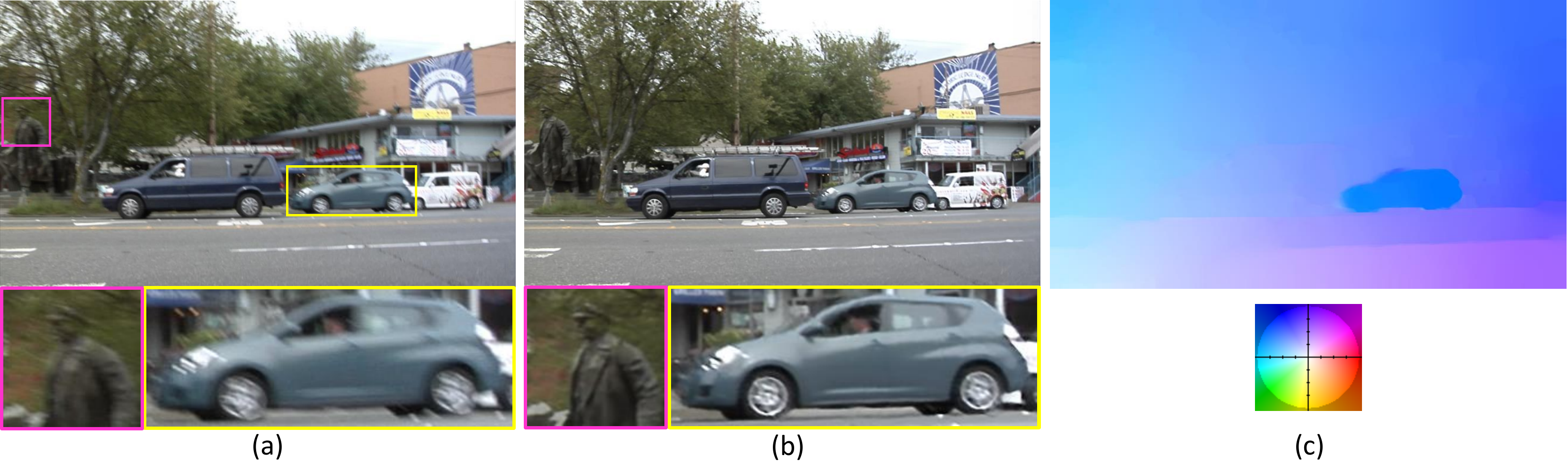}
	\end{center}
	\caption{(a) Blurry frame of video containing moving car. (b) Our deblurring result. (c) Our color coded optical flow.}
	\label{fig_intro}
\end{figure*}

However, previous approaches, which assume that the scene is static,
suffer from general blurs not only from camera shake but also from moving objects and depth variations
in a dynamic scene.
As parameterizing a spatially varying blur kernel in the dynamic scene is difficult with simple homography,
kernel estimation to handle dynamic scene becomes more challenging.
Therefore, several researchers have focused on restoring dynamic scenes,
which is mainly grouped into two approaches:
segmentation-based approach, and exemplar-based approach.

Segmentation-based deblurring approaches simultaneously estimate multiple motions, multiple kernels, 
and associated image segments.
Cho et al.~\cite{cho2007removing} proposed a method 
that segments images into multiple regions of homogeneous motions
and estimates the corresponding blur kernel as a one-dimensional Gaussian kernel.
Therefore, this method cannot handle complex motions of objects
and rotational motions of cameras that generate locally varying blurs.
Bar et al.~\cite{bar2007variational} proposed a layered model
and segmented images into two layers (foreground and background).
In addition, they estimated a linear blur kernel corresponding to a foreground layer.
Although this method can explicitly handle occluded regions using a layered model,
the kernel is limited to a one-dimensional box filter only, and only a static camera is allowed.
Wulff and Black~\cite{Wulff:ECCV:2014} extended the previous work of Bar et al.
They focused on estimating the parameters for both foreground and background motions.
However, the motions within each segment are only parameterized using the affine model,
and extending to multi-layered scenes is difficult because such task requires joint estimation of depth ordering of the layers.
In summary, segmentation-based approaches have the advantage of handling blurs by moving objects in dynamic scenes.
However, parameterizing the motions in each segment remains an issue~\cite{thkim_cvpr2014}.
That is, it fails to segment non-parametrically varying complex motions such as motions of people,
because doing so with the simple models used in~\cite{bar2007variational,Wulff:ECCV:2014} is difficult.

The works of Matsushita et al.~\cite{matsushita2006full} 
and Cho et al.~\cite{cho_siggraph2012} are typical exemplar-based approaches.
These works estimate latent frames by interpolating sharp patches,
that commonly exist in a long image sequence.
Therefore, these methods disregard accurate segmentation and deconvolution, enabling the emergence of ringing artifacts.
However, the former work cannot handle blurs by moving objects. Moreover, the latter one can only treat blurs by slightly moving objects in dynamic scenes
because it searches sharp patches of a blurry patch using globally parameterized kernel with homography.
Therefore, handling fast-moving objects, which have distinct motions from backgrounds, is difficult.
Moreover, it degrades mid-frequency textures, such as grasses and trees,
because this method does not use deconvolution with spatial priors
but use interpolation to restore latent frames, which renders smooth results.

To alleviate the problems in previous works, we propose a new generalized video deblurring method
that estimates latent frames without using global motion parametrization and segmentation.
We estimate bidirectional optical flows and use them to estimate pixel-wise varying kernels.
Therefore, we can naturally handle coexisting blurs by camera shake, moving objects with complex motions, 
and depth variations.
However, sharp frames are required to obtain accurate optical flows
because estimating flow fields is difficult between blurry images.
In addition, accurate optical flows are necessary to restore sharp frames.
This case is a typical chicken-and-egg problem,
and thus we simultaneously estimate both variables.
Therefore, we propose a new single energy model to solve our joint problem.
We also provide a framework and efficient techniques to optimize the model.
The result of our system is shown in Fig.\ref{fig_intro},
in which the moving car is successfully restored because accurate optical flows are jointly estimated.

By minimizing the proposed energy function,
we achieve significant improvements in numerous real challenging videos that other methods fail to do, as shown in Fig.\ref{fig_teaser}.
Furthermore, we estimate more accurate optical flows compared with
the state-of-the-art flow estimation method, that handles blurry images.
The performances are demonstrated in our extensive experiments.

\section{Generalized Video Deblurring}
Most conventional video deblurring methods suffer from the coexistence of various motion
blurs from dynamic scenes because the motions cannot be parameterized using global 
or segment-wise parameterization.
To handle general blurs, we propose a new energy model using pixel-wise
kernel estimation rather than global or segment-wise parameterization. 
As blind deblurring is a well-known ill-posed problem, 
our energy model not only consists of data and spatial regularization terms
but also a temporal term. 
The model is expressed as follows:
\begin{equation}
\begin{split}
\textbf{E} =  \textbf{E}_{data} + \textbf{E}_{temporal} + \textbf{E}_{spatial},
\end{split}
\label{equ_base}
\end{equation}
and the details of each term in~(\ref{equ_base}) are given in the following sections.

\subsection{Data Model based on Approximated Blur}
\begin{figure}[t]
	\begin{center}
		\includegraphics[width=0.9\linewidth]{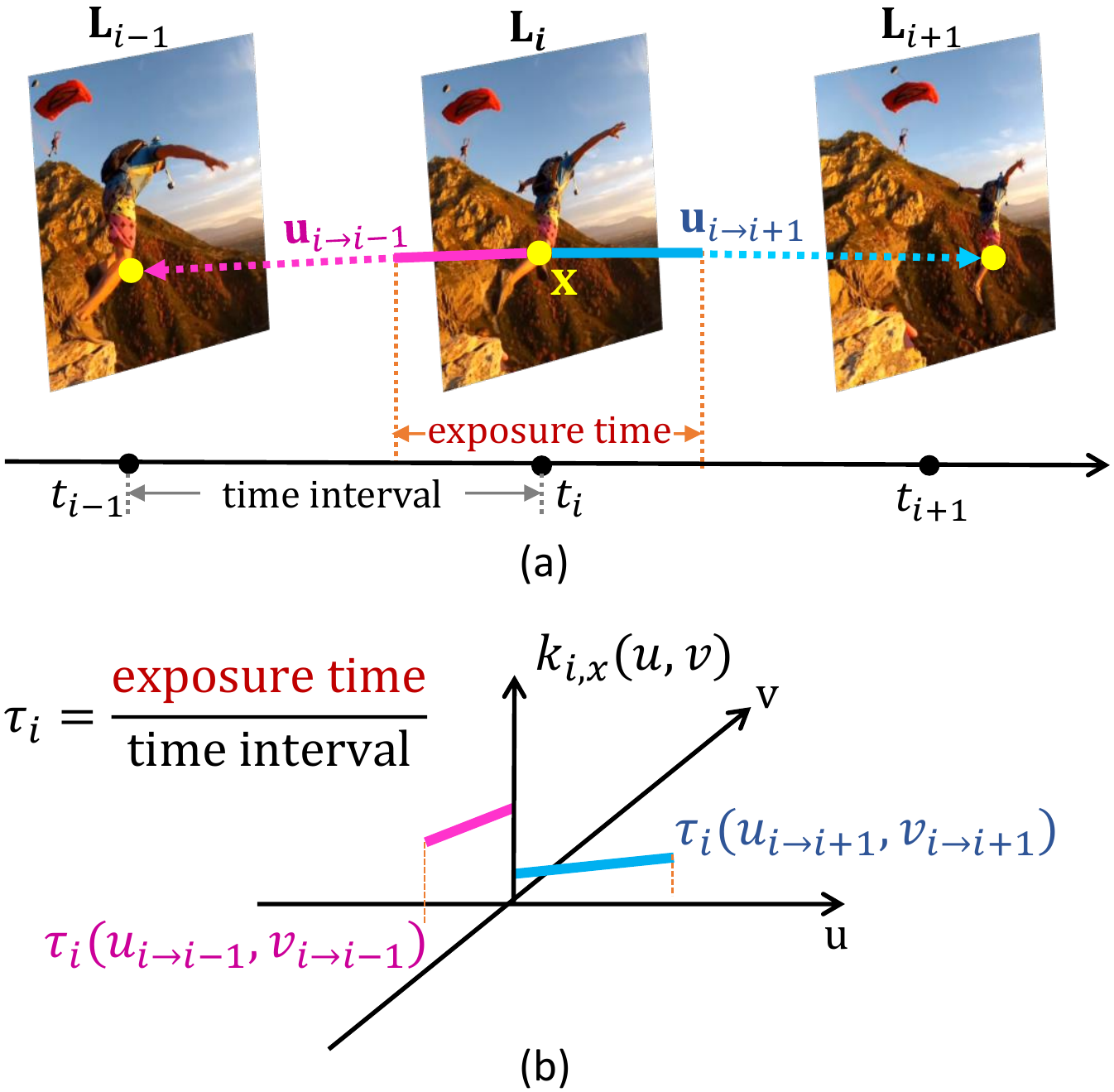}
	\end{center}
	\vspace{-3Ex}
	\caption{(a) Bidirectional optical flows. (b) Piece-wise linear blur kernel at pixel location $\textbf{x}$.}
	
	\label{fig_kernel}	
	\vspace{-2Ex}	
\end{figure}

In conventional works, the motion blurs of each frame are approximated using 
parametric models such as homographies and affine models~\cite{bar2007variational,cho_siggraph2012,li2010generating,Wulff:ECCV:2014}.
However, these kernel approximations are valid when motion blurs are parameterizable within an entire frame or segment.
Therefore, pixel-wise motion and kernel estimation are required to cope with general blurs.
We approximate the pixel-wise blur kernel using bidirectional optical flows,
in accordance with previous works~\cite{Dai:2008,thkim_cvpr2014,Portz:2012}.

Specifically, under an assumption that the velocity of the motion is constant between adjacent frames,
our blur model is expressed as follows:
\begin{equation}
\textbf{B}_i = \frac{1}{2\tau_i} \int_0^{\tau_i} H(\textbf{L}_i, t\cdot \textbf{u}_{i \rightarrow i+1}) + H(\textbf{L}_i, t\cdot \textbf{u}_{i \rightarrow i-1}) dt
\label{equ_blur_constraint},
\end{equation}
where $\textbf{u}_{i \rightarrow i+1}=(u_{i \rightarrow i+1}, v_{i \rightarrow i+1})$,
and $\textbf{u}_{i \rightarrow i-1}=(u_{i \rightarrow i-1}, v_{i \rightarrow i-1})$
denote bidirectional optical flows at frame $i$.
Blurry frame and latent frame are $\textbf{B}_i$ and $\textbf{L}_i$, respectively.
Camera duty cycle of the frame is $\tau_i$ and denotes relative exposure time~\cite{li2010generating}.
We define the image warping, $H(\textbf{L}_i, t\cdot \textbf{u}_{i \rightarrow i+1})$, which transforms the frame $\textbf{L}_i$ to $\textbf{L}_{i+t}$ when $0 \leq t \leq 1$, and $H(\textbf{L}_i, t\cdot \textbf{u}_{i \rightarrow i-1})$ transforms the frame $\textbf{L}_i$ to $\textbf{L}_{i-t}$. 
Our bi-directional optical flows,
duty cycle, and the corresponding piece-wise linear kernel
used in our blur model are illustrated in Fig.~\ref{fig_kernel}.

Although our blur kernel model is simple, 
our model can be justified because we treat video that has short exposure time to some extent.
Therefore, we approximate the kernel as piece-wise linear using bidirectional optical flows:
\begin{equation}
\begin{split}
&{k}_{i,x}(u,v) = \\
&\begin{cases}
\frac{\delta(u v_{i \rightarrow i+1} - v u_{i \rightarrow i+1})}{2\tau_i \|\textbf{u}_{i \rightarrow i+1}\|},&$if$~ u \in [0,\tau_i u_{i \rightarrow i+1}], v \in [0, \tau_i v_{i \rightarrow i+1}]\\
\frac{\delta(u v_{i \rightarrow i-1} - v u_{i \rightarrow i-1})}{2\tau_i \|\textbf{u}_{i \rightarrow i-1}\|},&$if$~ u \in (0,\tau_i u_{i \rightarrow i-1}], v \in (0, \tau_i v_{i \rightarrow i-1}]\\
0, &$otherwise.$
\end{cases}
\end{split},
\end{equation}
where ${k}_{i,x}(u,v)$ is the blur kernel using bidirectional optical flows at pixel location \textbf{x}, 
and $\delta$ denotes Kronecker delta.

\begin{figure}[t]
	\begin{center}
		\includegraphics[width=\linewidth]{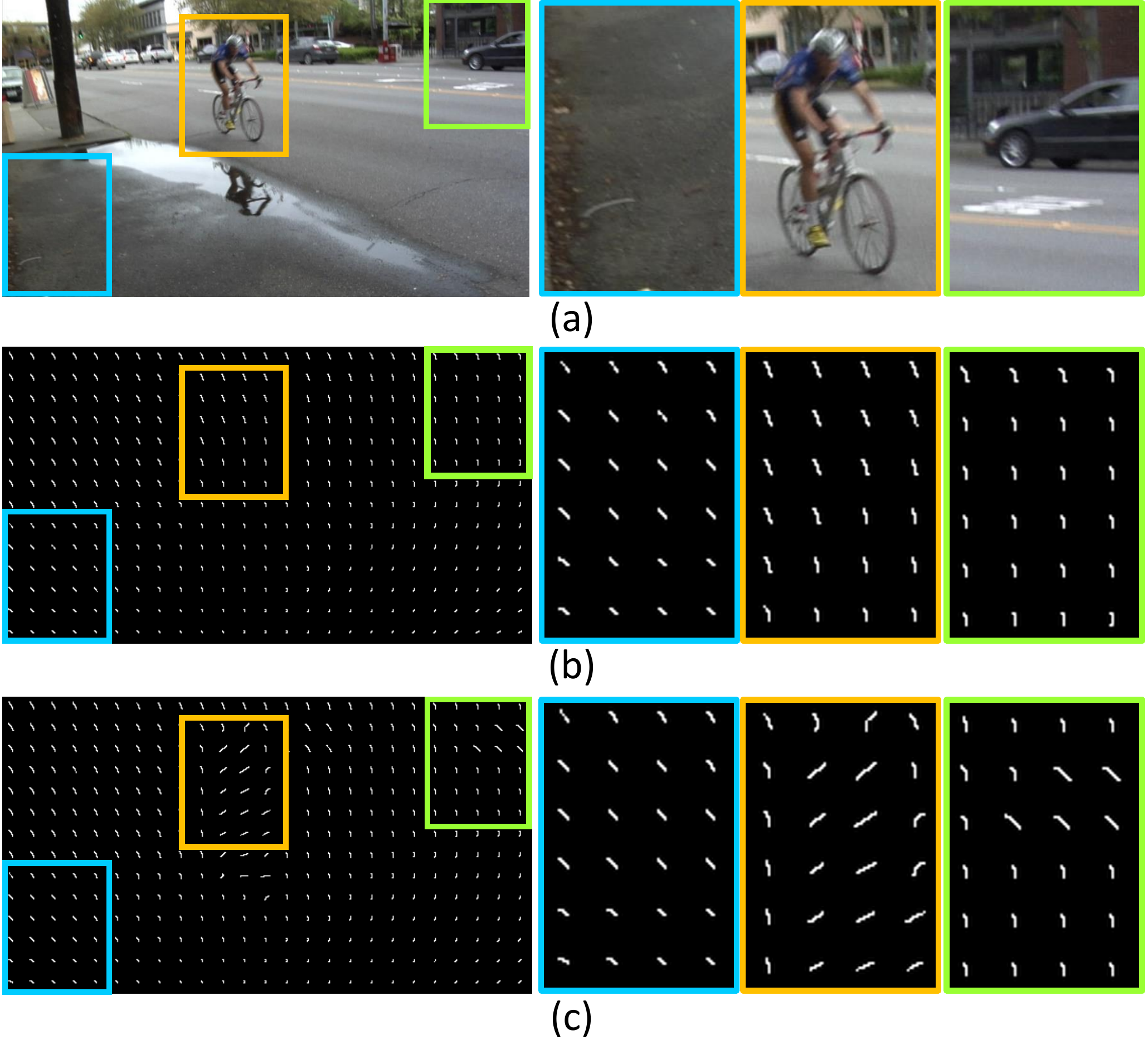}
	\end{center}
	\vspace{-4Ex}
	\caption{(a) Blurry frame of a video in dynamic scene. (b) Locally varying kernel using homography. (c) Our pixel-wise varying kernel using bidirectional optical flows.}
	\label{fig_kernel_difference}
	\vspace{-2Ex}	
\end{figure}

Using this pixel-wise kernel approximation,
we can easily manage multiple different blurs in a frame, unlike conventional methods.
The superiority of our kernel model is shown in Fig.~\ref{fig_kernel_difference}.
Our kernel model fits blurs from differently moving objects and camera shake much better
than the conventional homography-based model.

Therefore, we cast pixel-wise kernel estimation problem as an optical flows estimation problem.
Discretizing the constraint (\ref{equ_blur_constraint}) gives the following data term:
\begin{equation}
\begin{split}
&\textbf{E}_{data}({\textbf{L}, \textbf{u}, \textbf{B}}) =\\ &\lambda\sum_{i} \sum_{\partial_*}  \|\partial_*  \textbf{K}_i(\tau_i, \textbf{u}_{i \rightarrow i+1}, \textbf{u}_{i \rightarrow i-1}) \textbf{L}_i -\partial_* \textbf{B}_i \|^2,
\end{split}
\label{equ_data}
\end{equation}
where the row vector of blur kernel matrix $\textbf{K}_i$, corresponding to the blur kernel at pixel $\textbf{x}$, 
is the vector form of $k_{i,x}(.)$, and its elements are non-negative and their sum is equal to one. 
Linear operator $\partial_*$ denotes the Toeplitz matrices corresponding to the partial 
(e.g., horizontal and vertical) derivative filters.
Parameter $\lambda$ controls the weight of the data term,
and $\textbf{L}$, $\textbf{u}$, and $\textbf{B}$ denote the set of latent frames,
optical flows, and blurry frames, respectively.

\subsection{Temporal Coherence with Optical Flow Constraint}
Here, we determine that optical flows are required to estimate the pixel-wise blur kernel.
However, the proposed data term does not have conventional optical flow constraints 
such as brightness constancy or gradient constancy in (\ref{equ_data}).
In general, such constraints do not hold between two blurry frames.
Thus, Portz et al.~\cite{Portz:2012} proposed a method to apply flow constraints between blurry images. Based on the commutative law of shift invariance of kernels~\cite{Jin:cvpr2005},
the authors of~\cite{Portz:2012} convolved the approximated blur of each observed image to the other image 
and assumed constant brightness between them at matched points.
However, the commutativity property does not hold in theory when the kernel is not translation invariant.
Therefore, this approach only works when the motion is smooth enough.

To address this problem, we propose a new model that finds correspondences between two latent sharp images
to enable abrupt changes in motions and the corresponding kernels.
In using this model, we need not restrict our blur kernels to be shift invariant.
Our model is based on the conventional optical flow constraint between latent images,
that is, brightness constancy.
The formulation is expressed as follows:
\begin{equation}
\begin{split}
\textbf{E}_{temporal}(\textbf{L},\textbf{u}) = 
 \sum_{i} \sum_{n = -N}^{N} \mu_{n}   | \textbf{L}_i(\textbf{x}) - \textbf{L}_{i + n}(\textbf{x} + \textbf{u}_{i \rightarrow i+n})  |,
\end{split}
\label{equ_temporal}
\end{equation}
where $n$ denotes the index of neighboring frames at $i$. 
Constant parameter $\mu_{n}$ controls the weight of each term in the summation.
We apply the robust $L^1 $ norm to offer robustness against outliers and occlusions.
 
Notably, a major difference between the proposed model and the conventional optical flow estimation methods is that
our problem is a joint problem.
That is, the brightness of latent frames and optical flows need to be simultaneously estimated.
Therefore, our model simultaneously enforces the temporal coherence of latent frames 
and estimates the correspondences.

\subsection{Spatial Coherence}
To alleviate the difficulties of highly ill-posed deblurring and optical flow estimation problems,
several researchers have emphasized the importance of spatial regularization.
Therefore, we also enforce spatial coherence to penalize spatial fluctuations
while allowing discontinuities in both latent frames and flow fields.
We assume that spatial priors for latent frames and optical flows are independent.
They are expressed as follows:
\begin{equation}
\begin{split}
&\textbf{E}_{spatial}(\textbf{L},\textbf{u}) = \sum_i  | \nabla \textbf{L}_i | + \sum_{n=-N}^{N} g_i(\textbf{x}) | \nabla \textbf{u}_{i \rightarrow i+n} |.
\end{split}
\label{equ_spatial}
\end{equation}
The first term in (\ref{equ_spatial}) denotes the spatial regularization term for the latent frames.
Although more sparse $L^p$ norms (e.g., $p = 0.8$) fit the gradient statistics of natural sharp images better~\cite{Krishnan:2009,Krishnan:2011,Levin:PAMI2007}, 
we use conventional total variation (TV) based regularization~\cite{hu_cvpr2014_depthdeblur,thkim:2013,thkim_cvpr2014},
as TV is computationally less expensive. The second term denotes the spatial smoothness term for optical flows.
We adopt edge-map coupled TV-based regularization~\cite{thkim_iccv2013optical} 
to preserve discontinuities in the flow fields at edges.
Similar to~\cite{thkim_cvpr2014}, the edge-map is expressed as follows:
\begin{equation}
g_i(\textbf{x}) = \nu \exp(- (\frac {|\nabla \overline{\textbf{L}}_i|}{\sigma_I})^2),
\end{equation}
where $\nu$ controls the scale of the edge-map, parameter $\sigma_I$ controls the weight,
and $\overline{\textbf{L}}_i$ is an initial latent image in the iterative optimization framework.

\section{Optimization Framework}
 In the previous sections, we described the $\textbf{E}_{data}$, $\textbf{E}_{temporal}$, and $\textbf{E}_{spatial}$ terms.
When camera duty cycle $\tau_i$ is known, our final objective function becomes as follows:
\begin{equation}
\begin{split}
\min_{\textbf{L},\textbf{u}} &~ \lambda \sum_{i} \sum_{\partial_*}  \|\partial_*  \textbf{K}_i(\textbf{u}_{i \rightarrow i+1}, \textbf{u}_{i \rightarrow i-1}) \textbf{L}_i -\partial_* \textbf{B}_i \|^2 +\\
&\sum_{i} \sum_{n = -N}^{N} \mu_{n}  \cdot | \textbf{L}_i(\textbf{x}) - \textbf{L}_{i + n}(\textbf{x} + \textbf{u}_{i \rightarrow i+n})  | + \\
&\sum_{i} | \nabla \textbf{L}_i | + \sum_{n=-N}^{N} g_i(\textbf{x}) | \nabla \textbf{u}_{i \rightarrow i+n} |.
\end{split}
\label{equ_final}
\end{equation}
Unlike the work of Cho et al.~\cite{cho_siggraph2012}, which sequentially performs multi-phase approaches,
our model obtains a solution by minimizing a single objective function.
However, because of its non-convexity, our model is required to adopt practical optimization methods 
to obtain approximated solution.
Therefore, we divide the original problem into two sub-problems and use conventional iterative 
and alternating optimization techniques~\cite{Cho:2009,Wulff:ECCV:2014} to minimize the non-convex objective function.
In the following sections, 
we introduce efficient solvers and describe how to estimate unknowns $\textbf{L}$ and $\textbf{u}$, with one of them being fixed.
 
\subsection{Sharp Video Restoration}
While the optical flows $\textbf{u}$ are fixed, corresponding blur kernels are also fixed,
and our objective function in~(\ref{equ_final}) becomes convex with respect to $\textbf{L}$, and is expressed as follows:
\begin{equation}
\begin{split}
&\min_{\textbf{L}} ~\lambda  \sum_i \sum_{\partial_*} \|\partial_*  \textbf{K}_i \textbf{L}_i -\partial_* \textbf{B}_i \|^2 +\\
& \sum_i  \sum_{n = -N}^{N} \mu_{n} \cdot | \textbf{L}_i(\textbf{x}) - \textbf{L}_{i+n}(\textbf{x}+\textbf{u}_{i \rightarrow i+n}) | +  | \nabla \textbf{L}_i |.
\end{split} 
\label{equ_video_restoration}
\end{equation}
To obtain $\textbf{L}$, 
we adopt the conventional convex optimization method in~\cite{Chambolle:2011}, 
and derive the primal-dual update scheme as follows:
\begin{equation}\begin{cases}
\textbf{s}^{m+1}_i = \frac {\textbf{s}^m_i + \eta_L  \textbf{A} \textbf{L}_i^m}{\max(\textbf{1}, ~\text{abs}(\textbf{s}^m_i + \eta_L \textbf{A} \textbf{L}_i^m ) )}  \\
\textbf{q}^{m+1}_{i,n} = \frac {\textbf{q}^m_{i,n} + \eta_L \mu_n \textbf{D}_{i,n}
	\begin{bmatrix}
	\scriptsize\textbf{L}_i^m\\
	\scriptsize\textbf{L}_{i+n}^{m}
	\end{bmatrix}}
{\max(\textbf{1}, ~\text{abs}(\textbf{q}^m_{i,n} + \eta_L
	\mu_n \textbf{D}_{i,n}
	\begin{bmatrix}
	\scriptsize\textbf{L}_i^m\\
	\scriptsize\textbf{L}_{i+n}^{m}
	\end{bmatrix} )
	)}  \\

\begin{split}
&\textbf{L}_i^{m+1} = \arg \min_{\textbf{L}_i}  \lambda \sum_{\partial_*}  (\partial_* \textbf{K}_i\textbf{L}_i - \partial_*\textbf{B}_i)^2 +\\
& {\frac{(\textbf{L}_i - (\textbf{L}_i^m - \epsilon_L( \textbf{A}^T\textbf{s}_i^{m+1} + \sum_{n=-N}^N \mu_n{\textbf{D}_{i,n}}^T \textbf{q}_{i,n}^{m+1} )))^2}{2\epsilon_L}},
\end{split}
\end{cases}
\label{equ_update_L}
\end{equation}
where $m\geq0$ indicates the iteration number,
and, $\textbf{s}_i$ and $\textbf{q}_{i,n}$ denote the dual variables.
Parameters $\eta_L$ and $\epsilon_L$ denote the update steps.
A linear operator $\textbf{A}$ calculates the spatial difference between neighboring pixels,
and another operator $\textbf{D}_{i,n}$ calculates the temporal differences 
between $\textbf{L}_i(\textbf{x})$ and $\textbf{L}_{i+n}(\textbf{x}+\textbf{u}_{i \rightarrow i+n})$.
To update the primal variable and obtain $\textbf{L}_i^{m+1}$ in~(\ref{equ_update_L}), 
we apply the conjugate gradient method to optimize the quadratic function.

\subsection{Optical Flows Estimation}
While the latent frames $\textbf{L}$ are fixed, temporal coherence term $\textbf{E}_{temporal}$ becomes convex
but the data term $\textbf{E}_{data}$ remains non-convex.
Therefore, we define a non-convex fidelity function $\rho(.)$ as follows:
\begin{equation}
\begin{split}
\rho(\textbf{x},\textbf{u})= \lambda &\sum_i \sum_{\partial_*}  \| \partial_* \textbf{K}_i(\textbf{u}_{i \rightarrow i+1}, \textbf{u}_{i \rightarrow i-1}) \textbf{L}_i - \partial_* \textbf{B}_i \|^2 +\\
 & \sum_i \sum_{n=-N}^N \mu_n \cdot | \textbf{L}_i(\textbf{x}) - \textbf{L}_{i+n}(\textbf{x}+\textbf{u}_{i \rightarrow i+n}) |.
\end{split}
\end{equation}
To find the optimized values of optical flows $\textbf{u}$, 
we first convexify the non-convex function $\rho(.)$ by applying the first-order Taylor expansion.
Similar to~\cite{thkim_cvpr2014}, we linearize the function near an initial $\textbf{u}_0$ in the iterative process as follows:
\begin{equation}
\rho(\textbf{x},\textbf{u}) \approx \rho(\textbf{x},\textbf{u}_0) + \nabla \rho(\textbf{x},\textbf{u}_0)^T(\textbf{u}-\textbf{u}_0).
\end{equation}
Therefore, our approximated convex function for optical flows estimation is expressed as follows:
\begin{equation}
\begin{split}
\min_{\textbf{u}} \rho(\textbf{x},\textbf{u}_0) +  \nabla \rho(\textbf{x}, \textbf{u}_0)^T(\textbf{u}-\textbf{u}_0) + \sum_i\sum_{n=-N}^{N} g_i(\textbf{x}) |\nabla \textbf{u}_{i\rightarrow i+n}|.
\end{split}
\label{equ_convex_data}
\end{equation}
~Next, we apply the convex optimization technique in~\cite{Chambolle:2011}
to the approximated convex function (\ref{equ_convex_data}),
and the primal-dual update process is expressed as follows:
\begin{equation}\begin{cases}
\textbf{p}_{i,n}^{m+1} = \frac {\textbf{p}_{i,n}^m + \eta_u (\textbf{G}_i\textbf{A}) \textbf{u}_{i \rightarrow i+n}^m}{\max(\textbf{1},  ~\text{abs}(\textbf{p}_{i,n}^m + \eta_u (\textbf{G}_i\textbf{A}) \textbf{u}_{i \rightarrow i+n}^m))}  \\
\textbf{u}_{i \rightarrow i+n}^{m+1} = (\textbf{u}_{i \rightarrow i+n}^m - \epsilon_u(\textbf{G}_i\textbf{A})^T \textbf{p}_{i,n}^{m+1}) - \epsilon_u \nabla_{i,n}\rho(\textbf{x},\textbf{u}_0),
\end{cases}
\label{equ_update_motion}
\end{equation}
where $\textbf{p}_{i,n}$ denotes the dual variable of $\textbf{u}_{i \rightarrow i + n}$ on the vector space
and the diagonal matrix $\textbf{G}_i$ is the weighting matrix denoted as $\textbf{G}_i= \text{diag}(g_i(\textbf{x}))$.
Parameters $\eta_u$ and $\epsilon_u$ denote the update steps
and $\nabla_{i,n}\rho(\textbf{x},\textbf{u}_0)$ means $\frac{\partial \rho(\textbf{x},\textbf{u})}{\partial \textbf{u}_{i \rightarrow i + n}} | \textbf{u}_0$.

\section{Implementation Details}
\begin{figure}[t]
	\begin{center}
		\includegraphics[width=\linewidth]{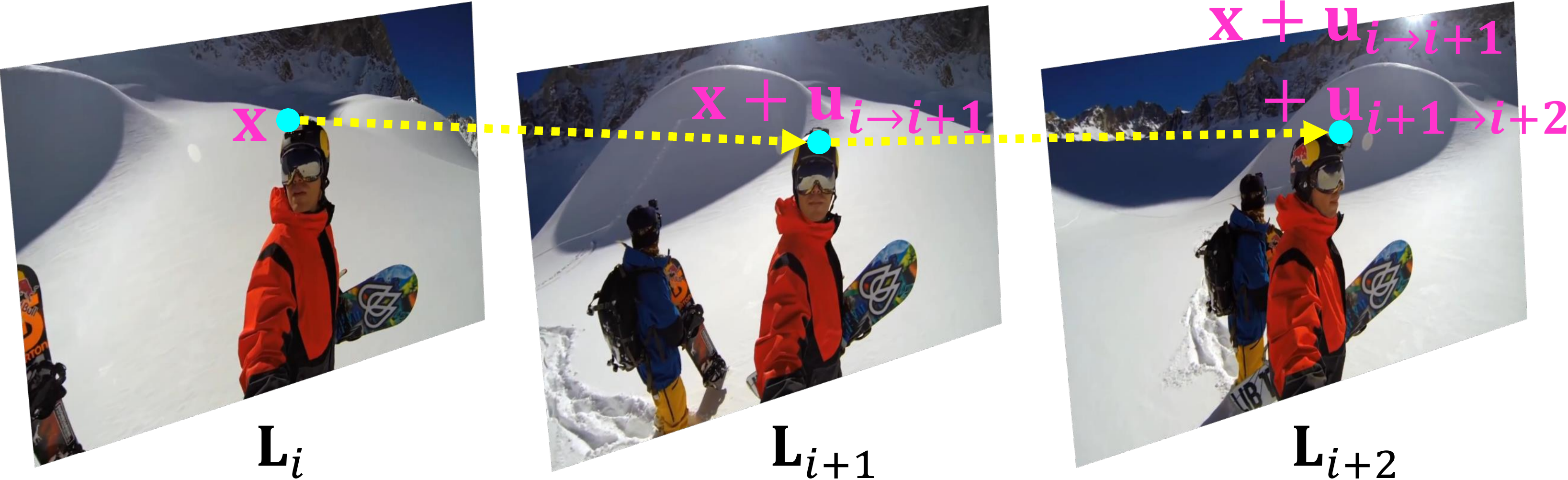}
	\end{center}
	\caption{Temporally consistent optical flows over three frames.}
	\label{fig_corr}
	\vspace{-3Ex}	
\end{figure}

To handle large blurs and guide fast convergence, we implement our algorithm on the traditional coarse-to-fine framework with empirically determined parameters.
We use $\lambda = 250$ for our most experiments,
and other parameters are determined as $\mu_n = \lambda$, $\nu = 0.08\lambda$, $\sigma_I = \frac{25}{255}$, and $N = 2$.
In the coarse-to-fine framework, we build image pyramid with 17 levels 
for a high-definition(1280x720) video, the scale factor is 0.9, 
and use bi-cubic interpolation to propagate both the optical flows and latent frames to the next pyramid level. 

Moreover, to reduce the number of unknowns in optical flows, 
we only estimate $\textbf{u}_{i \rightarrow i+1 }$ and $\textbf{u}_{i \rightarrow i-1 }$.
We approximate $\textbf{u}_{i \rightarrow i+2 }$ using $\textbf{u}_{i \rightarrow i+1 }$ and $\textbf{u}_{i+1 \rightarrow i+2 }$. 
For example, it satisfies, $\textbf{u}_{i \rightarrow i+2} = \textbf{u}_{i \rightarrow i+1} + \textbf{u}_{i+1 \rightarrow i+2}$, as illustrated in Fig.~\ref{fig_corr}, and we can easily apply this for $n \neq 1$.

The overall process of our algorithm is in Algorithm \ref{algorithm_overall}.
Further details on estimating the duty cycle $\tau_i$
and post-processing step that reduces artifacts are given below.

\begin{algorithm}[h]
	\footnotesize
	\caption{Overview of the proposed method}
	\begin{algorithmic}[1]
		\renewcommand{\algorithmicrequire}{\textbf{Input:}}
		\renewcommand{\algorithmicensure}{\textbf{Output:}}
		\REQUIRE Blurry frames $\textbf{B}$
		\ENSURE  Latent frames $\textbf{L}$ and optical flows $\textbf{u}$
		\STATE {Initialize duty cycle ${\tau_i}$ and optical flows \textbf{u}. (Sec. 4.1)}
		\STATE {Build image pyramid.}
		\STATE {Restore \textit{sharp video} with fixed \textbf{u}. (Sec. 3.1)}
		\STATE {Estimate \textit{optical flows} with fixed \textbf{L}. (Sec. 3.2)}
		\STATE {Detect occlusion and perform post-processing. (Sec 4.2)}
		\STATE {Propagate variables to the next pyramid level if exists.}
		\STATE {Repeat steps 3-6 from coarse to fine pyramid level.}
	\end{algorithmic} \label{algorithm_overall}	
\end{algorithm}

\subsection{Duty Cycle Estimation}
In this study, we assume that the camera duty cycle $\tau_i$ is known for every frame.
We can obtain the duty cyle from public SDK, when we use Kinect to capture RGB videos.
However, when we conduct deblurring with conventional data sets, which do not provide exposure information,
we apply the technique proposed in~\cite{cho_siggraph2012} to estimate the duty cycle.
Contrary to the original method in~\cite{cho_siggraph2012},
we use optical flows instead of homographies to obtain initially approximated blur kernels.
Therefore, we first estimate flow fields from blurry images with~\cite{wedel2009improved}, which runs in near real-time.
We then use them as initial flows and approximate the kernels to estimate the duty cycle.

\subsection{Occlusion Detection and Refinement}
Our piece-wise linear kernel naturally results in approximation error
and it causes problems such as ringing artifacts.
Moreover, our data model in (\ref{equ_data}), 
and temporal coherence model in (\ref{equ_temporal}) are invalid at occluded regions.

To reduce such artifacts from kernel errors and occlusions, we use spatio-temporal filtering as a post-processing:
\begin{equation}
\textbf{L}^{m+1}_i(\textbf{x})^ = \frac{1}{Z(\textbf{x})}\sum_{n=-N}^N \sum_\textbf{y} w_{i,n}(\textbf{x},\textbf{y})\cdot \textbf{L}_{i+n}^m(\textbf{y}),
\label{equ_spatio_temporal_filter}
\end{equation}
where \textbf{y} denotes a pixel in the 3x3 neighboring patch at location $(\textbf{x}+\textbf{u}_{i \rightarrow i+n})$
and $Z$ is the normalization factor (e.g. ${Z(\textbf{x})} = \sum_{n=-N}^N \sum_\textbf{y} w_{i,n}(\textbf{x},\textbf{y})$).
Notably, we enable $n=0$ in (\ref{equ_spatio_temporal_filter}) for spatial filtering.
Our occlusion-aware weight $w_{i,n}$ is defined as follows:
\begin{equation}
w_{i,n}(\textbf{x},\textbf{y}) = o_{i,n}(\textbf{x},\textbf{y}) \cdot \exp(- \frac{  \| P_i(\textbf{x}) - P_{i+n}(\textbf{y}) \|^2 } {2\sigma_w^2} ),
\end{equation}
where occlusion state $o_{i,n}(\textbf{x},\textbf{y}) \in \{0, 0.5, 1\} $ is determined using the method proposed in~\cite{thkim_iccv2013optical}.
The 5x5 patch $P_i(\textbf{x})$ is centered at $\textbf{x}$ in frame $i$.
The similarity control parameter $\sigma_w$ is fixed as $\sigma_w = 25/255$.

\section{Experimental Results}

\begin{figure*}[t]
	\begin{center}
		\includegraphics[width=\linewidth]{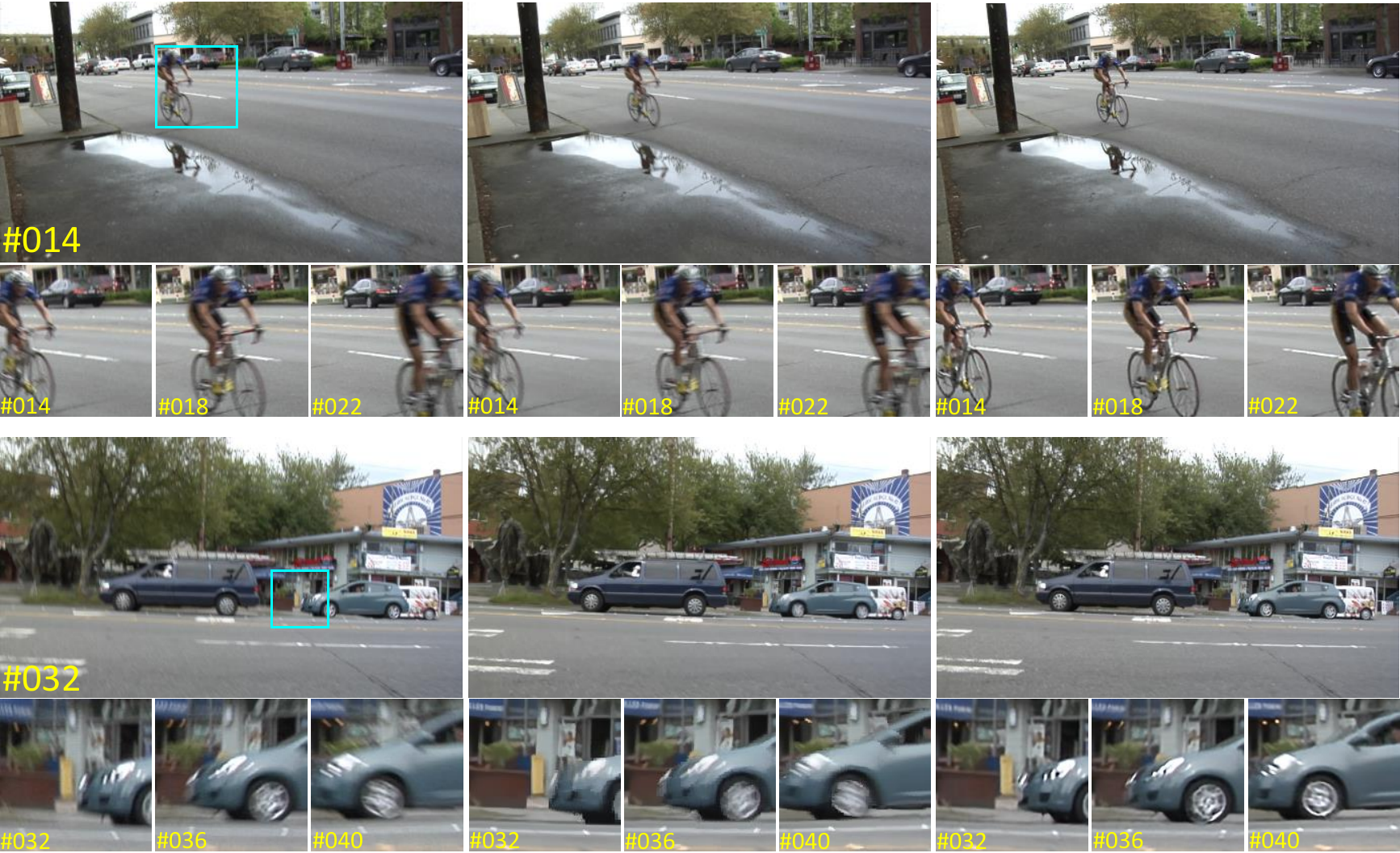}
	\end{center}
		\vspace{-3Ex}
	\caption{\textbf{Left to right:} Blurry frames of dynamic scenes, deblurring results of \cite{cho_siggraph2012}, and our results.}
	\label{fig_comp_exemplar1}
	
	\begin{center}
		\includegraphics[width=\linewidth]{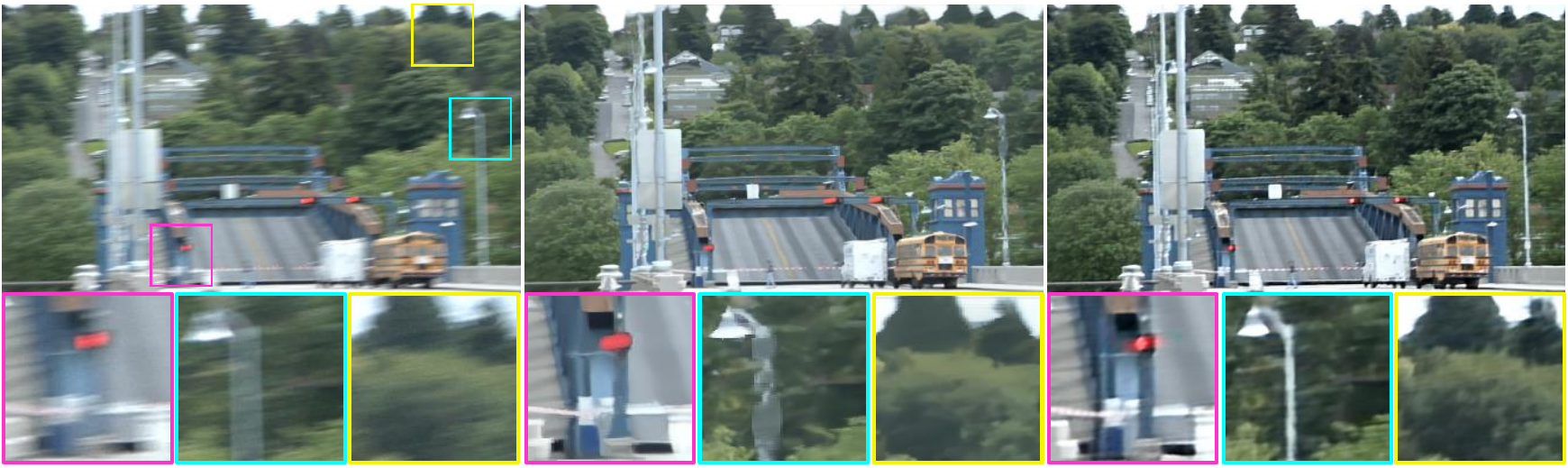}
	\end{center}
		\vspace{-3Ex}
	\caption{\textbf{Left to right:} Blurry frame, deblurring result of \cite{cho_siggraph2012}, and ours.}
	\label{fig_comp_exemplar2}

	%
	\begin{center}
		\includegraphics[width=\linewidth]{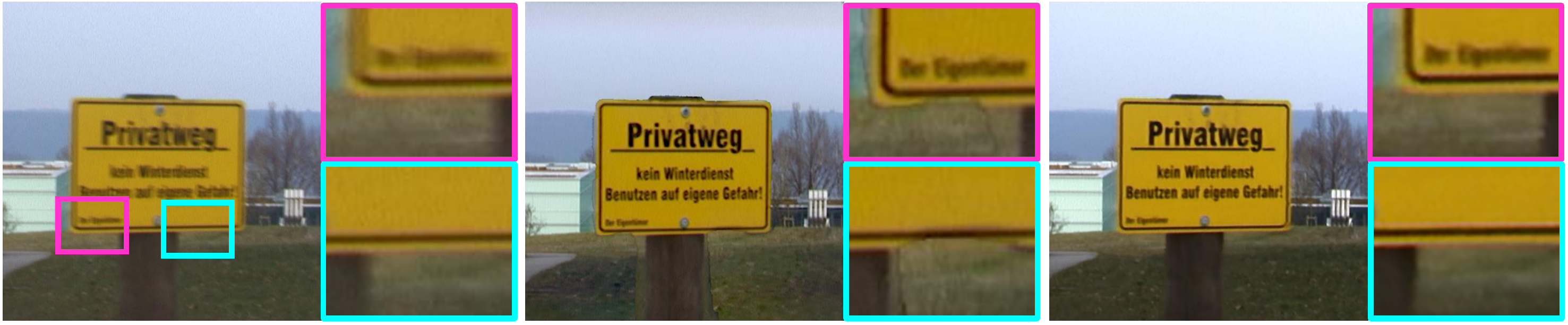}
	\end{center}
		\vspace{-3Ex}
	\caption{Comparison with segmentation-based approach. \textbf{Left to right:} Blurry frame, result of \cite{Wulff:ECCV:2014}, and ours.}
	\label{fig_com_segmentation}	
\end{figure*}

In what follows, we demonstrate the superiority of the proposed method.
(For more results, see the supplementary video.)

First, we compare our deblurring results with 
those of the state-of-the art exemplar based method \cite{cho_siggraph2012} 
with the videos used in \cite{cho_siggraph2012}.
As shown in Fig. \ref{fig_comp_exemplar1}, the captured scenes are dynamic and contain multiple moving objects.
The method \cite{cho_siggraph2012} fails in restoring the moving objects,
because the object motions are large and distinct from the backgrounds.
By contrast, our results show better performances in deblurring moving objects and backgrounds.
This exemplar-based approach also fails in handling large blurs, as shown in Fig. \ref{fig_comp_exemplar2},
as the initially estimated homographies in the largely blurred images are inaccurate.
Moreover, this approach renders excessively smooth results for mid-frequency textures such as trees,
as the method is based on interpolation without spatial prior for latent frames.

Next, we compare our method with the state-of-the-art segmentation-based approach \cite{Wulff:ECCV:2014}.
In Fig.~\ref{fig_com_segmentation}, the captured scene is a bi-layer and used in \cite{Wulff:ECCV:2014}.
Although the bi-layer scene is a good example to verify the performance of the layered model,
inaccurate segmentation near the boundaries causes serious artifacts in the restored frame.
By contrast, our method does not depend on accurate segmentation
and thus restores the boundaries much better than the layered model.

\begin{figure}[t]
	\begin{center}
		\includegraphics[width=\linewidth]{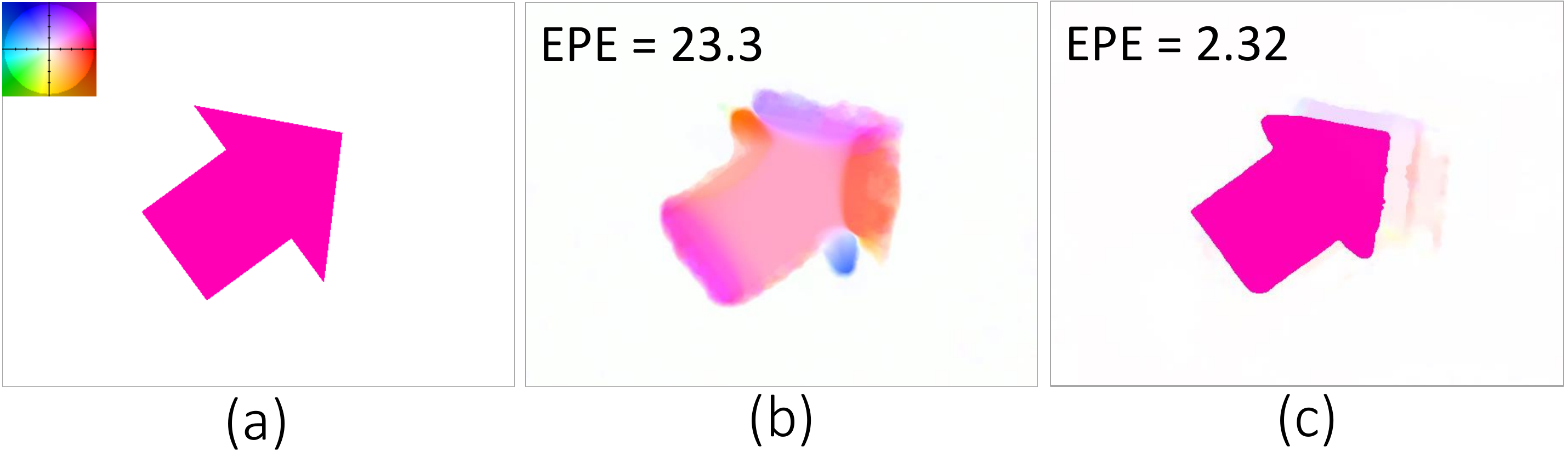}
	\end{center}
	\caption{EPE denotes average end point error. (a) Color coded ground truth optical flow between blurry images. (b) Optical flow estimation result of \cite{Portz:2012}. (c) Our result.}
	\label{fig_comp_optical_flow}
	%
	
	\vspace{5Ex}
	\begin{center}
		\includegraphics[width=\linewidth]{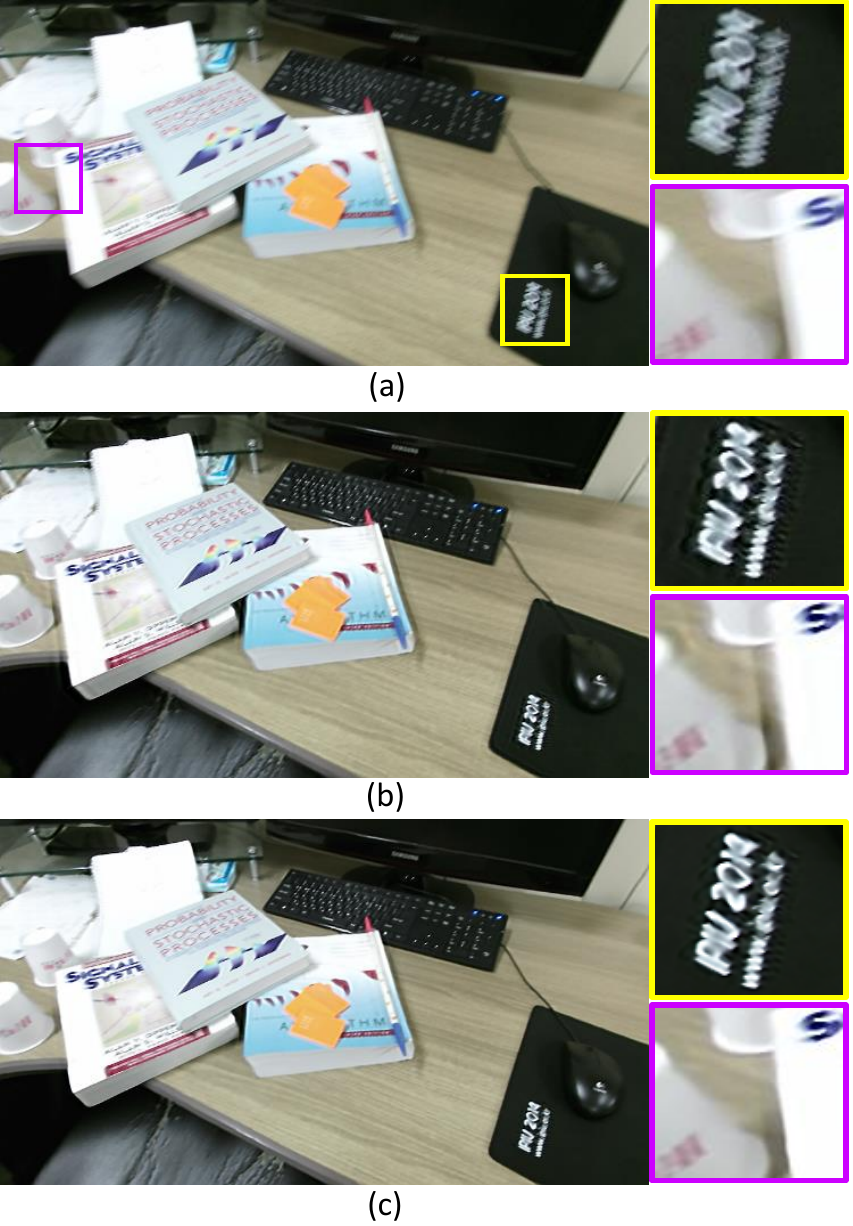}
	\end{center}
	\caption{(a) Real blurry frame of a video. (b) Our deblurring result without using $\textbf{E}_{temporal}$. (c) Our deblurring result with $\textbf{E}_{temporal}$.}
	\label{fig_kinect_ipiu}

\end{figure}

In Fig.~\ref{fig_comp_optical_flow}, we quantitatively compare the optical flow accuracies with \cite{Portz:2012}
on synthetic blurry images.
Although \cite{Portz:2012} proposed to handle blurry images in optical flow estimation,
its assumption does not hold in motion boundaries, which are very important for deblurring.
Therefore, their optical flow is inaccurate in the motion boundaries of moving objects.
However, our model enables abrupt changes of motions and thus performs better than the previous model.

Moreover, we show the deblurring results
with and without using the temporal coherence term in (\ref{equ_temporal}),
and verify that our temporal coherence model significantly reduces ringinig artifacts near the edges in Fig.~\ref{fig_kinect_ipiu}.

\begin{figure}[t]
	\begin{center}
		\includegraphics[width=\linewidth]{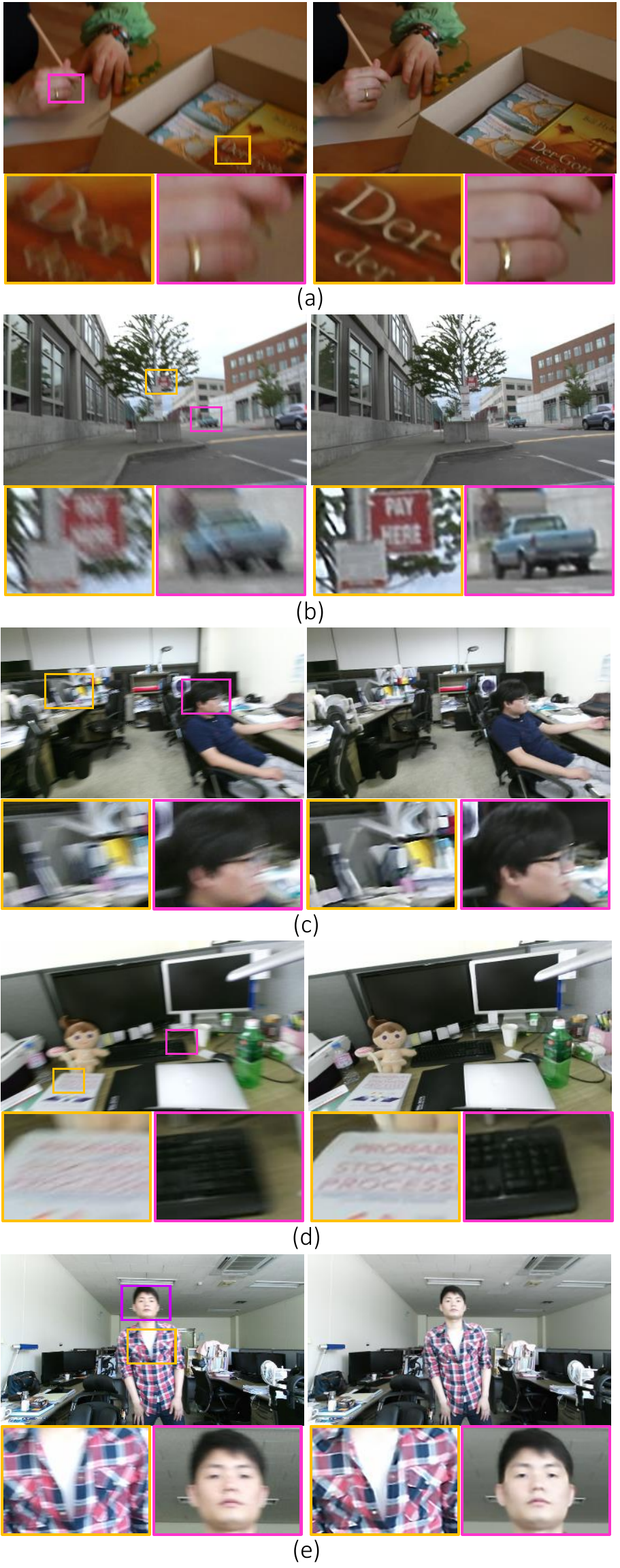}
	\end{center}
	\vspace{-3Ex}
	\caption{\textbf{Left to right:} Numerous real blurry frames and our deblurring results. (a)-(b) Data sets used in \cite{cho_siggraph2012}. (c)-(e) Captured RGB data sets using kinect.}
	\label{fig_comp_self}	
\end{figure}

Other deblurring results from numerous real videos are shown in Fig.~\ref{fig_comp_self}.
Notably, our model successfully restores the face which has highly non-uniform blurs 
because the person moves rotationally (Fig.~\ref{fig_comp_self}(e)). 


\section{Conclusions}
In this study, we introduced a novel method that removes general blurs in dynamic scenes,
which conventional methods fail to do.
By estimating a pixel-wise kernel using optical flows, we handled general blurs.
Thus, we proposed a new energy model that estimates optical flows and latent frames, jointly.

We also provided a framework and efficient solvers to minimize the energy function
and achieved significant improvements in removing general blurs in dynamic scenes.

%

\clearpage
\section*{Acknowledgments}
This research was supported in part by the MKE (The Ministry of Knowledge Economy), Korea and Microsoft Research, under IT/SW Creative research program supervised by the NIPA (National IT Industry Promotion Agency) (NIPA-2013-H0503-13-1041), and in part by the National Research Foundation of Korea (NRF) grant funded by the Ministry of Science, ICT \& Future Planning (MSIP) (No. 2009-0083495)
{\small
\bibliographystyle{ieee}
\bibliography{VD}
}

\end{document}